\documentclass[sigconf]{acmart}

\copyrightyear{2021} 
\acmYear{2021} 
\setcopyright{acmcopyright}\acmConference[MM '21]{Proceedings of the 29th ACM International Conference on Multimedia}{October 20--24, 2021}{Virtual Event, China}
\acmBooktitle{Proceedings of the 29th ACM International Conference on Multimedia (MM '21), October 20--24, 2021, Virtual Event, China}
\acmPrice{15.00}
\acmDOI{XX.XXXX/XXXXXXX.XXXXXXX}
\acmISBN{978-1-4503-8651-7/21/10}

\usepackage{amsmath}
\usepackage{xspace}
\usepackage{amsmath}
\usepackage{balance}
\usepackage{makecell}
\usepackage{multirow}

\DeclareMathOperator*{\argmax}{arg\,max}

\makeatletter
\DeclareRobustCommand\onedot{\futurelet\@let@token\@onedot}
\def\@onedot{\ifx\@let@token.\else.\null\fi\xspace}

\def\eg{\emph{e.g}\onedot}

\def\etal{\emph{et al}\onedot}

\usepackage[misc]{ifsym} 
\usepackage{graphicx}
\usepackage{relsize}
\newcommand\smalldownarrow{\mathrel{\scalebox{0.8}[0.5]{$\downarrow$}}}
\newcolumntype{P}[1]{>{\centering\arraybackslash}p{#1}}
\makeatother

\begin{document}
\fancyhead{}

\title{Combining Attention with Flow for Person Image Synthesis}

\author{Yurui Ren}
\affiliation{%
  \institution{School of Electronic and Computer Engineering, Peking University}
  \country{}
  }
\email{yrren@pku.edu.cn}

\author{Yubo Wu}
\affiliation{%
  \institution{School of Electronic and Computer Engineering, Peking University}
  \country{}
}

\author{Thomas H. Li}
\affiliation{%
 \institution{AIIT, Peking University \\ ITRDIT, Peking University}
 \country{}
 }

\author{Shan Liu}
\affiliation{%
  \institution{Tencent America}
  \country{}
  }
  \email{shanl@tencent.com}

\author{Ge Li}
\authornote{Corresponding Author}
\affiliation{%
  \institution{School of Electronic and Computer Engineering, Peking University}
  \country{}
  }
\email{geli@ece.pku.edu.cn}



\renewcommand{\shortauthors}{Trovato and Tobin, et al.}

\begin{abstract}
  Pose-guided person image synthesis aims to synthesize person images by transforming reference images into target poses.
  In this paper, we observe that the commonly used spatial transformation blocks have complementary advantages.
  We propose a novel model by combining the attention operation with the flow-based operation. 
  Our model not only takes the advantage of the attention operation to generate accurate target structures but also uses the flow-based operation to sample realistic source textures.
  Both objective and subjective experiments demonstrate the superiority of our model.
  Meanwhile, comprehensive ablation studies verify our hypotheses and show the efficacy of the proposed modules.
  Besides, additional experiments on the portrait image editing task demonstrate the versatility of the proposed combination.
\end{abstract}




\maketitle
\begin{figure}[t]
\begin{center}
\includegraphics[width=0.9\linewidth]{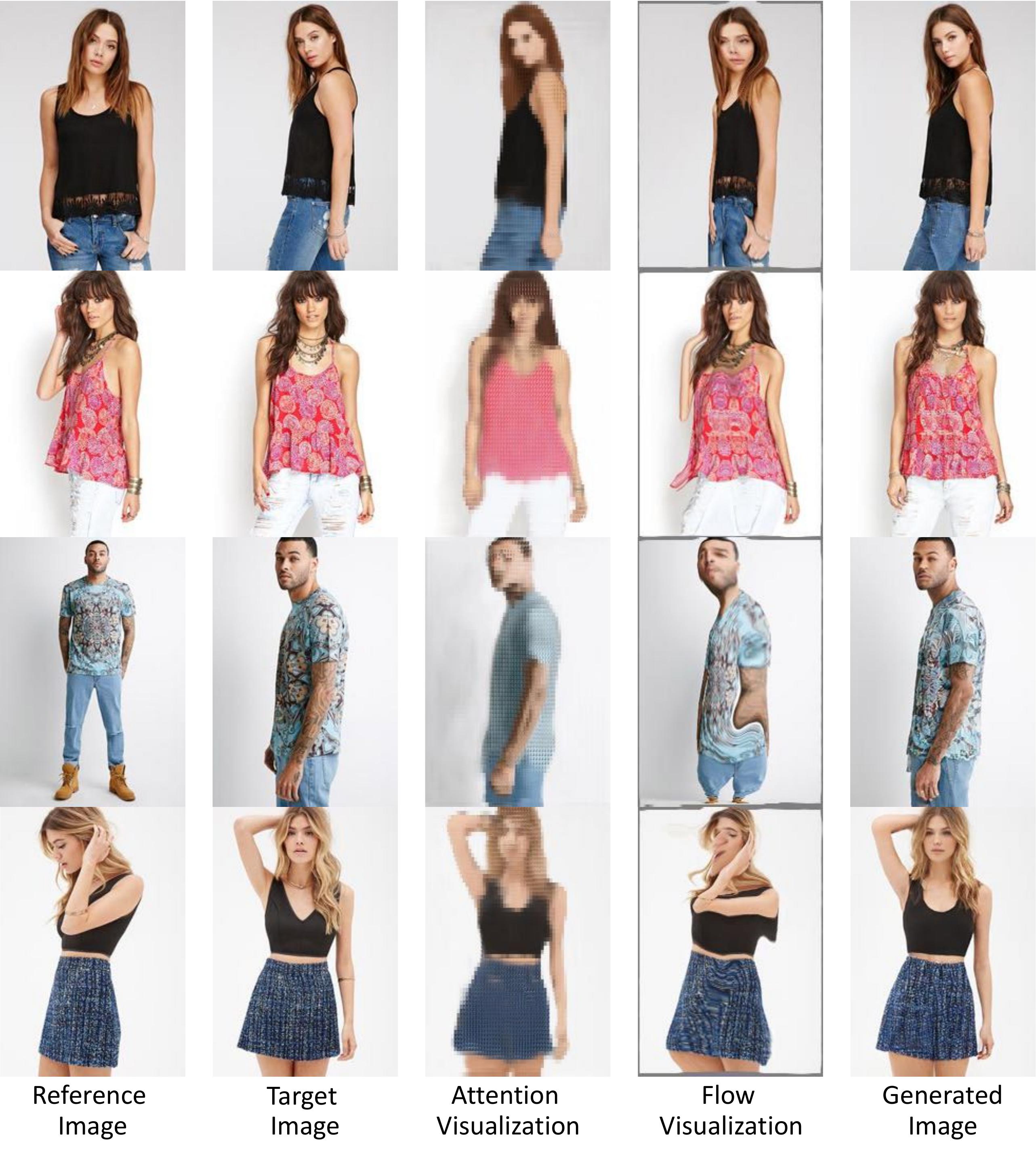}
\end{center}
\caption{The generated examples and the corresponding visualizations of the deformations obtained by the attention and flow-based operations. Our model takes the complementary advantages of flow and attention.}
\label{fig:intro}
\Description{The generated examples of our model.}
\end{figure}

\section{Introduction}
Being able to synthesize person images by transforming the poses of given persons is an important task with a large variety of applications. Industries such as electronic commerce, 
virtual reality, film production, and next-generation communication require such algorithms to generate content.
In many cases, however, these requirements are achieved by graphic technologies with precise control over the image rendering. Editing images in this way needs professionals to build fine-grained 3D models for each scene. This complex and tedious process prevents ordinary users from these algorithms and meanwhile increases the costs of the generated contents.

Recently, advances in computer vision fields have made tremendous progress in generating realistic images~\cite{goodfellow2014generative,brock2018large,karras2019style,karras2020analyzing}. Some algorithms~\cite{ma2017pose,siarohin2018deformable,ren2020deep,men2020controllable} are proposed to automatically synthesize person images from references using learning-based methods. Formally, the pose-guided person image synthesis task aims to synthesize person images by transforming the poses of reference images according to the given modifications while preserving the reference identities. Examples are provided in Fig.~\ref{fig:intro}. 
It can be seen that the reference and target images in this task have clear mapping relationships: targets are the spatial transformation versions of the reference images.
Therefore, this task can be tackled by reassembling the references in the spatial domain.
However, Convolutional Neural Networks (CNNs) lack the abilities to enable efficient spatial transformation~\cite{goodfellow2016deep,vaswani2017attention}. Convolutional operations are building blocks that process one local neighborhood at a time. To model long-term dependencies, stacks of convolutional operations are required. Realistic source textures will be ``wash away'' during these operations, which results in over-smoothed images. Therefore, a fundamental challenge for this task is to design efficient spatial transformation blocks to reassemble the reference images.

The attention operation has been proved as an effective method to extract non-local dependencies~\cite{vaswani2017attention,wang2018non,zhang2019self}. The response of a target query is calculated as the weighted sum of source features.
By using this operation, each target feature directly communicates with all source features. Thus, targets can be expected to sample specific source features by increasing the weights of the corresponding regions and refusing the other features. 
However, to generate images with realistic textures, each output position should only sample a very local region of sources. 
This requires the attention correlation matrix to be a sparse matrix to reject all the unsampled features, which is extremely difficult for the standard attention operation. 
Meanwhile, this position-irreverent operation cannot maintain the patterns of the reference images (\eg logos).

Another efficient spatial transformation operation is the flow-based operation. 
This operation warps the source information by predicting 2D coordinate offsets specifying the sampling positions. Different from the attention operation, the flow-based operation can obtain photo-realistic textures since it sampling a very local source patch for each target position. However, it is hard for the networks to obtain stable gradients from the flow-based operation because each output feature is only related to a local and indeterminate source patch. This phenomenon will hinder the models to extract accurate motions, which is more evident when complex deformations and severe occlusions are observed. 

\begin{figure*}[t]
\begin{center}
\includegraphics[width=1\linewidth]{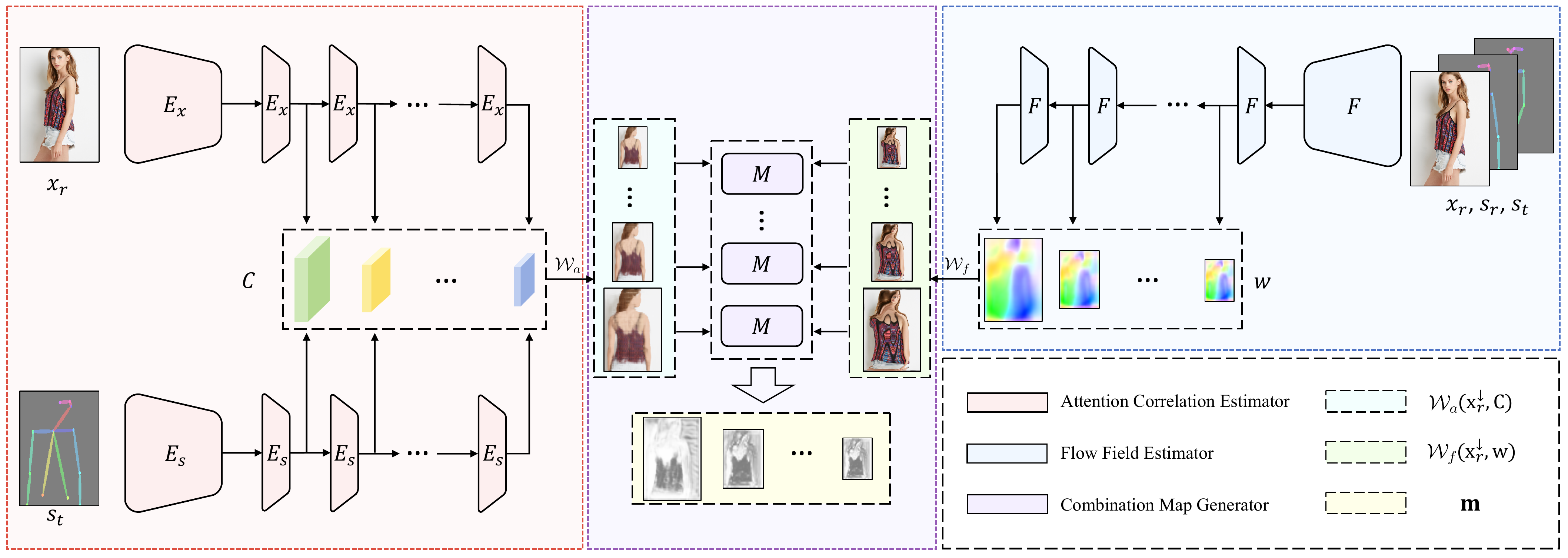}
\end{center}
\caption{The architecture of the deformation estimation module. The deformations are estimated by the attention correlation estimator and the flow field estimator. Then, the combination maps are generated using the corresponding warped images. Our model estimates multi-scale deformations to transform both global and local contexts. }
\label{fig:deformation}
\Description{The architecture of the deformation estimation module.}
\end{figure*}

Observing the complementary advantages of these two operations, in this paper, we propose a novel model by combining the attention operation with the flow-based operation.
The architecture of the proposed model is shown in Fig.~\ref{fig:deformation} and Fig.~\ref{fig:generation}. 
Specifically, a \textit{Deformation Estimation Module} is first designed to extract the deformations between the reference images and the desired targets. 
Two types of deformations are estimated: the correlation matrices for the attention operation and the flow fields for the flow-based operation.
Then, we generate a combination map that is responsible for selecting the better deformation  between the correlation matrices and the flow-fields for each target position.
Finally, an \textit{Image Synthesis Module} is employed to synthesize the target images by reassembling the source feature maps according to the estimated deformations and combination maps. 

We compare the proposed model with several state-of-the-art methods. The experiment results show the superiority of our model. Meanwhile, comprehensive ablation studies are conducted to verify the hypothesis and show the efficacy of the proposed modules. 
Besides, we further apply our model to tackle the portrait image editing task. 
We show that the proposed model can achieve intuitive portrait image control by modifying the poses and expressions of reference images according to the provided modifications. 
The main contributions of our paper can be summarized as
\begin{itemize}
  \item We propose a novel model for person image synthesis by combining the attention operation with the flow-based operation. 
  Taking the complementary advantages of these operations, our model can synthesize images with not only accurate structures but also realistic details.

  \item We demonstrate the versatility of the proposed model by further extending it to tackle the portrait image editing task. 
  Experiments show that our model can synthesize portrait images with accurate movements.

\end{itemize}

\section{Related Work}

Recently, deep neural networks are starting to produce visually compelling images conditioned on certain user specifications like segmentation and edge map.~\cite{isola2017image,choi2018stargan,huang2018multimodal,liu2019few,yu2019multi}.
The pose-guided person image synthesis task is a highly active topic in this field, where the target images are synthesized by rendering the corresponding skeletons with the appearance of the reference images.
Ma \etal~\cite{ma2017pose} tackle this task by proposing a coarse-to-fine framework. Their framework first synthesizes coarse images with accurate poses and then refine the results by adding vivid textures in an adversarial way.
Some follow-up works~\cite{ma2018disentangled,esser2018variational} manage to disentangle the poses and appearance of the reference images to improve the results. 
However, these methods use 1D embeddings to represent the appearance, which hinders the generation of complex textures.
Men \etal~\cite{men2020controllable} alleviate this problem by extracting appearance from different semantics separately. 
The extracted embeddings are then injected into the feature maps of target skeletons to generate the final results.
Instead of using the same embeddings for all target positions, Zhang \etal~\cite{zhang2021pise} propose to inject the style of each semantic part into the corresponding target semantic regions. 
With the representative appearance embeddings, these methods can generate images with realistic textures.
However, they cannot maintain the patterns of the reference images.
Meanwhile, these methods rely on accurate human parsing maps. 
The performance may be vulnerable to parsing errors.

Some other methods tackle this task by proposing efficient spatial transformation modules~\cite{siarohin2018deformable,wang2019few,liu2019liquid,zablotskaia2019dwnet,ren2020deep,tang2021structure}.
Siarohin \etal~\cite{siarohin2018deformable} introduce deformable skip connections to spatially transform the source neural textures with a set of affine transformations. 
This method relieves the spatial misalignment caused by pose difference and achieves good results.
However, it requires one to predefine a set of transformation components, which limits the application.
Zhu \etal~\cite{zhu2019progressive} use cascaded attention blocks to transfer the source information progressively. 
This method can generate accurate structures for target images.
But complex textures are smoothed during multiple transfers, which leads to the performance decline.
Li \etal~\cite{li2019dense} propose a flow-based method for this task.
To estimate accurate deformations, they generate flow field labels with additional 3D human reconstruction methods.
However, the performance of this model is limited by the accuracy of the 3D reconstruction model.
Han \etal~\cite{han2019clothflow} propose a cascaded flow estimator to predict flow fields in an unsupervised manner. 
However, this method warps the sources at the pixel level, which means that additional refinement networks are required to fill the holes caused by occlusions.
Ren \etal~\cite{ren2020deep} propose a global-flow local-attention framework to reassemble the input image at the feature level.
Tang \etal~\cite{tang2021structure} further improve the results by proposing a structure-aware person image synthesis method that predicts flow fields of different body semantics separately.
Recently, some methods~\cite{zhang2020cross,zhou2021cocosnet} achieve spatial transformation by using the attention operation. 
The correspondences between the sources and targets are extracted by leaning attention correlation metrics.
However, the attention operation cannot maintain the spatial distributions of the reference images, which hinders the methods remonstrating complex textures. 

\section{Our Approach}

We propose a novel model for the pose-guided person image synthesis task. 
The motivation of our model comes from the observation that the commonly used spatial transformation operations have complementary advantages.
As shown in Fig.~\ref{fig:intro}, the flow-based operation can extract vivid source textures by assigning a very local patch for each target position. 
However, it lacks the ability to capture complex deformations between sources and targets.
On the contrary, the attention operation can extract accurate deformations and synthesize targets with reasonable structures. However, it cannot maintain the source textures. 
Therefore, by combining these two operations, our model can synthesize images with not only accurate global structures but also realistic local details.
In the following, we first provide the details of our deformation estimation module (Sec.~\ref{sec:deformation_estimation}). Then, the image synthesis module is introduced for the target image synthesis (Sec.~\ref{sec:image_generation}). 
Finally, we explain the training functions (Sec.~\ref{sec:loss_function}).
Please note that we describe the model warping source features at a single scale for the simplicity of the discussion.
Our model can be extended by warping multi-scale source features to transform both global and local contexts.

\subsection{Deformation Estimation Module}
\label{sec:deformation_estimation}

A fundamental challenge of the pose-guided person image synthesis task is to accurately reassemble the source information according to the provided modifications. 
This requires one to estimate the correspondence between the reference image $\mathbf{I}_r$ and the desired target image $\mathbf{I}_t$.
We deal with this task using a deformation estimation module.
The architecture of this module is shown in Fig.~\ref{fig:deformation}. 
It consists of three parts: the attention correlation estimator, the flow field estimator, and the combination map generator. 

\noindent
\textbf{The Attention Correlation Estimator} is responsible for calculating the attention correlation matrix $\mathbf{C}$ that contains the correlations of all queries to all keys. 
This estimator first encodes the reference image $\mathbf{x}_r$ and the target skeleton $\mathbf{s}_t$ to feature maps using encoders $E_x$ and $E_s$ respectively. 
\begin{equation}
\begin{aligned}
  \mathbf{k} &= E_x(\mathbf{x}_r) \\
  \mathbf{q} &= E_s(\mathbf{s}_t)  
\end{aligned}
\end{equation}
where $\mathbf{k} \in \mathbb{R}^{H \times W \times K}$ and $\mathbf{q} \in \mathbb{R}^{H \times W \times K}$ are the extracted feature maps representing keys and queries, respectively. $H$ and $W$ denote the spatial size of the feature map and $K$ is the number of feature channels.
Then, the correlation matrix $\mathbf{C} \in \mathbb{R}^{HW\times HW}$ is obtained as
\begin{equation}
  \mathbf{C}^{i,j} = \frac{\text{exp}(\alpha \beta^{i,j})}{\sum_{i \in \Omega}\text{exp}(\alpha \beta^{i,j})}, \text{where } \beta^{i,j}={\mathbf{k}^i}^T \mathbf{q}^j
\end{equation}
where $\Omega$ is the coordinate set of the feature maps. 
Symbols $\mathbf{k}^i \in \mathbb{R}^K$ and $\mathbf{q}^j \in \mathbb{R}^K$ denote the 
feature located in $i$ of the reference feature map $\mathbf{k}$ and the feature located in $j$ of the skeleton feature map $\mathbf{q}$ respectively.
Coefficient $\alpha$ is used to control the sharpness of the softmax operation. In this paper, we set $\alpha=100$ for all experiments.
With matrix $\mathbf{C}$, we can calculate the attention results $\mathbf{o} \in \mathbb{R}^{H\times W\times K'}$ by weighted sum the source inputs $\mathbf{x} \in \mathbb{R}^{H\times W\times K'}$.
\begin{equation}
   \mathbf{o}^j = \mathcal{W}_a(\mathbf{x}, \mathbf{C})^j = \sum_{i \in \Omega} \mathbf{C}^{i, j}\mathbf{x}^i
\end{equation} 
where $\mathcal{W}_a(*,*)$ indicates the attention warping operation.
By using the attention operation, the source information can be integrated into desired target positions according to matrix $\mathbf{C}$.

\begin{figure}[t]
\begin{center}
\includegraphics[width=1\linewidth]{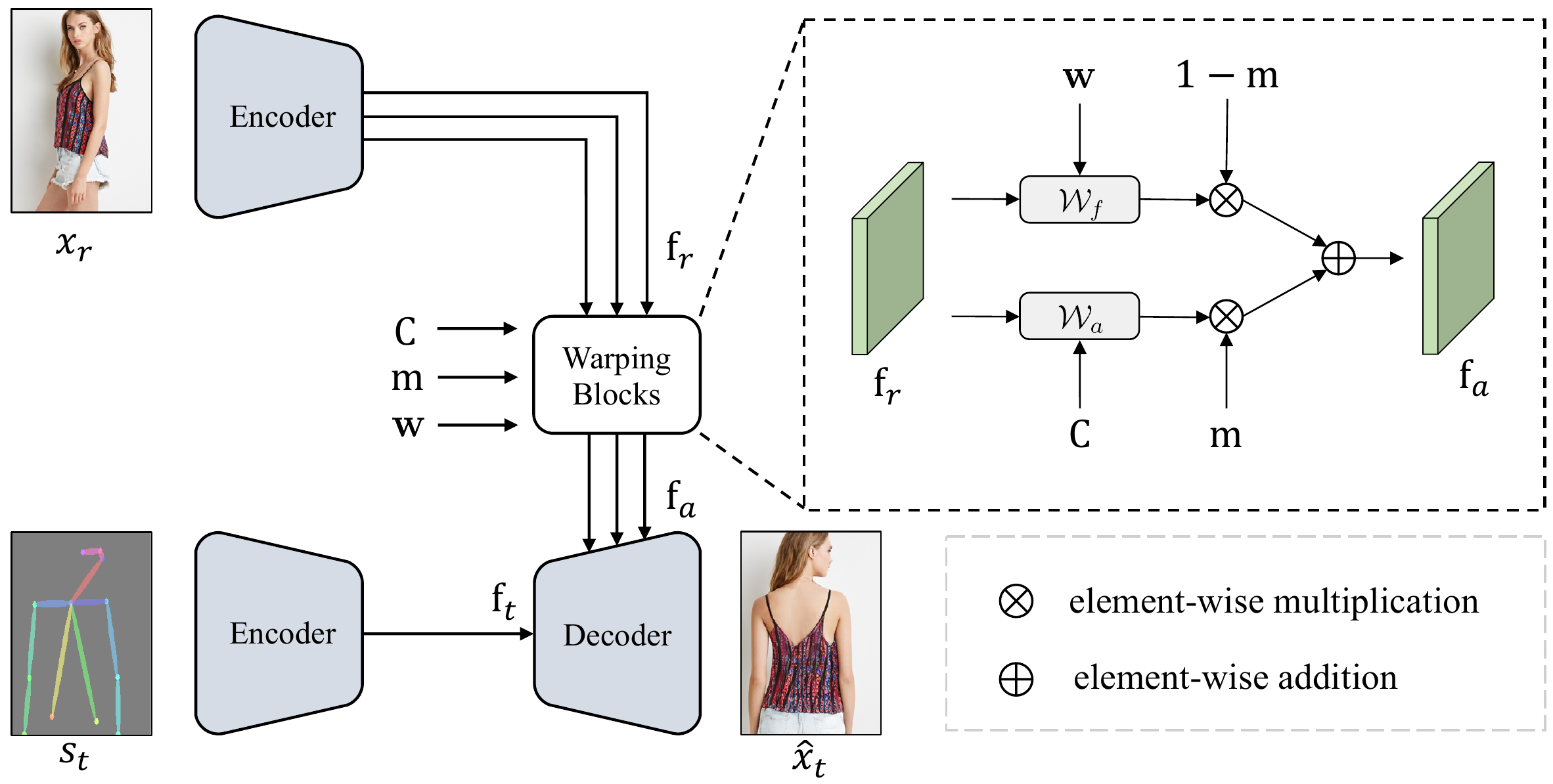}
\end{center}
\caption{The architecture of the image synthesis module. 
The feature maps of the reference images are warped using the estimated deformations. Then, the network generates the final images by rendering the target skeletons with aligned neural textures.
}
\label{fig:generation}
\Description{The architecture of the image generation module.}
\end{figure}

\noindent
\textbf{The Flow Field Estimator} is responsible for estimating flow fields that contain the relative movements between the sources and targets. 
Different from the attention operation, the flow-based operation forces the correlation matrices to be sparse matrices by sampling a local source region for each output. 
Therefore, this operation can help with reconstructing vivid source textures. 
To be specific, the estimator $F$ takes the reference image $\mathbf{x}_r$, the reference skeleton $\mathbf{s}_r$, and the target skeleton $\mathbf{s}_t$ as the inputs. The flow fields $\mathbf{w} \in \mathbb{R}^{H\times W\times 2}$ are generated by analyzing the difference between the reference images and the desired targets.
\begin{equation}
  \mathbf{w} = F(\mathbf{s}_t, \mathbf{s}_r, \mathbf{x}_r)
\end{equation}
We design $F$ with an auto-encoder structure. It first extracts features from the inputs and then decodes them into flow fields according to the extracted features. Several skip-connections are used for leveraging both local and global contexts. 
After obtaining $\mathbf{w}$, the output results of the flow-based operation can be obtained by 
\begin{equation}
  \mathbf{o} = \mathcal{W}_f(\mathbf{x}, \mathbf{w}) 
\end{equation}
where $\mathcal{W}_f(*,*)$ indicates the flow-based warping operation which samples the input $\mathbf{x}$ with flow fields $\mathbf{w}$ using the Bilinear interpolation method.

\noindent
\textbf{The Combination Map Generator} predicts the combination maps to select from the warping results of the attention operation and the warping results of the flow-based operation. 
As mentioned before, the attention operation and the flow-based operation have complementary advantages.
Therefore, reasonably combining their results is important for exploiting their strengths and thus improve the quality of the final images.
Here, we tackle this task by generating combination maps $\mathbf{m}\in \mathbb{R}^{H\times W\times 1}$ using a content-aware combination map generator $M$. 
\begin{equation}
  \mathbf{m} = M(\mathcal{W}_a(\mathbf{x}_r^{\smalldownarrow}, \mathbf{C}), \mathcal{W}_f(\mathbf{x}_r^{\smalldownarrow}, \mathbf{w}))
\end{equation}
where $\mathbf{x}_r^{\smalldownarrow}$ is the resized image of the original reference image $\mathbf{x}_r$.
We design generator $M$ with several residual blocks. The Sigmoid activation function is used as the non-linear function of the output layer. The combination maps $\mathbf{m}$ have continuous values between $0$ and $1$. 
With the deformations $\mathbf{C}$, $\mathbf{w}$ and the combination maps $\mathbf{m}$, we can generate the final images by spatially transforming the source textures.


\subsection{Image Synthesis Module}
\label{sec:image_generation}
The image synthesis module $G$ is used to synthesize the final images by rendering target skeletons with reference textures.
The architecture of this module is shown in Fig.~\ref{fig:generation}. Specifically, this module takes $\mathbf{x}_r$, $\mathbf{s}_t$, $\mathbf{C}$, $\mathbf{w}$, and $\mathbf{m}$ as inputs and synthesizes the predicted images $\hat{\mathbf{x}}_t$.
\begin{equation}
  \hat{\mathbf{x}}_t = G(\mathbf{x}_r, \mathbf{s}_t, \mathbf{C}, \mathbf{w}, \mathbf{m})
\end{equation}
The warping block in module $G$ is responsible for reassembling the reference neural textures according to the estimated deformations. Let $\mathbf{f}_r$ denotes the feature map extracted from the reference image $\mathbf{x}_r$. This block generates the aligned feature map $\mathbf{f}_a$ by first warping $\mathbf{f}_r$ with both the attention correlation matrix $\mathbf{C}$ and the flow field $\mathbf{w}$ and then combining the warped results with the combination map $\mathbf{m}$.
\begin{equation}
  \mathbf{f}_a = \mathbf{m}\otimes \mathcal{W}_a(\mathbf{f}_r, \mathbf{C}) + (1-\mathbf{m}) \otimes \mathcal{W}_f(\mathbf{f}_r, \mathbf{w})
\end{equation}
where $\otimes$ denotes the element-wise multiplication over the spatial domain.
After obtaining $\mathbf{f}_a$, the target image $\hat{\mathbf{x}}_t$ is generated by adding vivid neural textures to the feature map $\mathbf{f}_t$ extracted from the target skeleton $\mathbf{s}_t$
\begin{equation}
  \mathbf{f}_o = \mathbf{f}_t + \mathbf{f}_a
\end{equation}
where $\mathbf{f}_o$ is the output feature map containing both target semantics and reference textures. We further decode $\mathbf{f}_o$ to synthesize $\hat{\mathbf{x}}_t$

\subsection{Training Losses}
\label{sec:loss_function}

The proposed model is trained with several loss functions that fulfill special tasks. These loss functions can be divided into two categories: losses for accurate deformations and losses for realistic images.

\noindent
\textbf{Losses for Accurate Deformations.} 
Estimating accurate deformations between sources and targets is crucial for generating realistic images.
However, using losses at the end of the network (\eg reconstruction loss) cannot guarantee that the model learns meaningful deformations. Therefore, several losses are designed to directly constrain the estimated deformations.
For the attention correlation matrix $\mathbf{C}$, we calculate the $\ell_1$ distance between the warped image and the target image.
\begin{equation}
  \mathcal{L}_{attn} = \lVert \mathcal{W}_a(\mathbf{x}_r^{\smalldownarrow}, \mathbf{C}), \mathbf{x}_t^{\smalldownarrow} \rVert_1
\end{equation}
This attention loss $\mathcal{L}_{attn}$ encourages the correlation matrix $\mathbf{C}$ containing meaningful deformations to reduce the reconstruction errors.
To constrain the flow field $\mathbf{w}$, we employ the sampling correctness loss $\mathcal{L}_{flow}$ and the regularization loss $\mathcal{L}_{regu}$ proposed in paper~\cite{ren2020deep}.
The sampling correctness loss calculates the normalized cosine similarity between the warped reference feature and the ground-truth target feature. We use VGG-19 to extract the corresponding features $\mathbf{v}_r$, $\mathbf{v}_t$ from images $\mathbf{x}_r$, $\mathbf{x}_t$. This loss is defined as
\begin{equation}
  \mathcal{L}_{flow} = \frac{1}{N} \sum_{l\in\Omega}exp[-\frac{\mu\big(\mathcal{W}_f(\mathbf{v}_r, \mathbf{w})^l, \mathbf{v}_t^l\big)}{\mu\big(\mathbf{v}_r^{l_{max}}, \mathbf{v}_t^l\big)}]
\end{equation}
where $\mu(*)$ indicates the cosine similarity operation. Coordinate set $\Omega$ contains all $N$ positions in the feature maps.
For each position $l$, this loss first calculates the similarity between the warped reference feature $\mathcal{W}_f(\mathbf{v}_r, \mathbf{w})^l$ and the ground-truth target feature $\mathbf{v}_t^l$. Then, the similarity is normalized by $\mu(\mathbf{v}_r^{l'}, \mathbf{v}_t^l)$ to avoid the bias brought by occlusion, where $\mathbf{v}_r^{l_{max}}$ is the most similar feature of $\mathbf{v}_t^l$ in $\mathbf{v}_r$.
\begin{equation}
  l_{max} = \argmax_{i\in \Omega}\mu(\mathbf{v}_r^{i}, \mathbf{v}_t^l)
\end{equation}
The regularization loss $\mathcal{L}_{regu}$ is used to extract the spatial correlations of the flow fields $\mathbf{w}$. 
It assumes that each local deformation estimated by $\mathbf{w}$ should be an affine transformation. 
Let $\mathbf{R}_l=\left[{\begin{array}{cccc}x_1&x_2&...&x_{n\times n}\\
y_1&y_2&...&y_{n \times n}\\
1&1&...&1
\end{array}}\right]$ be the reference coordinates of a local $n\times n$ patch centered at location $l$. The coordinates of the sampling points can be calculated using the corresponding flow field patch $\mathbf{w}_l$ with $\mathbf{S}_l=\mathbf{R}_l+\mathbf{w}^*_l$, where $\mathbf{w}^*_l$ is the homogeneous coordinates of $\mathbf{w}_l$. This loss assumes a linear relationship between $\mathbf{R}_l$ and $\mathbf{S}_l$ and calculates the least-square error.
\begin{equation}
  \mathcal{L}_{regu} = \sum_{l \in \Omega} \left\lVert \mathbf{R}_l - \hat{\mathbf{A}}_l \mathbf{S}_l \right\lVert_2^2
\end{equation}
where the metric $\hat{\mathbf{A}}_l$ is the least-square solution of the linear equation $\mathbf{R}_l = \mathbf{A}_l\mathbf{S}_l$. It can be calculated as
\begin{equation}
    \hat{\mathbf{A}}_l = \mathbf{R}_l\mathbf{S}_l^T(\mathbf{S}_l\mathbf{S}_l^T)^{-1}
\end{equation}

\noindent
\textbf{Losses for Realistic Images.} 
After obtaining accurate deformations, our model synthesizes the final images with the warped features. Several losses are designed to obtain realistic images. The perceptual loss proposed in paper~\cite{johnson2016perceptual} is used to calculate the reconstruction error between the predicted image $\hat{\mathbf{x}}_t$ and the ground-truth image $\mathbf{x}_t$.
\begin{equation}
  \mathcal{L}_{perc} = \sum_i \lVert \phi_i(\mathbf{x}_t), \phi_i(\hat{\mathbf{x}}_t) \rVert_1 
\end{equation}
where $\phi_i$ denotes the $i$-th activation map of the VGG-19 network. 
Besides, a face reconstruction loss is used for generating natural faces.
This loss calculates the difference between features of the cropped faces.
\begin{equation}
  \mathcal{L}_{face} = \sum_i \lVert \phi_i\big(C_{face}(\mathbf{x}_t)\big), \phi_i\big(C_{face}(\hat{\mathbf{x}}_t)\big) \rVert_1 
\end{equation}
where $C_{face}(*)$ is the face cropping function. 
To generate vivid details, we use a style loss to calculate the statistical error between the activation maps.
\begin{equation}
  \mathcal{L}_{style} = \sum_j \lVert G^\phi_j(\mathbf{x}_t), G^\phi_j(\hat{\mathbf{x}}_t) \rVert_1 
\end{equation}
where $G^\phi_j$ represents the Gram matrix of activation map $\phi_j$. 
In addition to the VGG-based losses, a generative adversarial loss is employed to mimic the distribution of ground-truth images. 
\begin{equation}
  \mathcal{L}_{adv} = \mathbb{E}(1-D(\hat{\mathbf{x}}_t) + \mathbb{E}(D(\mathbf{x}_t))
\end{equation}
where $D$ is the discriminator. We use the following overall loss to train our model.
\begin{equation}
\begin{aligned}
  \mathcal{L} &= \lambda_{attn}\mathcal{L}_{attn}+\lambda_{flow}\mathcal{L}_{flow}+\lambda_{regu}\mathcal{L}_{regu} \\
  &+\lambda_{perc}\mathcal{L}_{perc} +\lambda_{face}\mathcal{L}_{face}  + \lambda_{style}\mathcal{L}_{style} \\
  &+ \lambda_{adv}\mathcal{L}_{adv}
\end{aligned}
\end{equation}

\section{Experiment}
\noindent
\textbf{Dataset.} 
The \textit{In-shop Clothes Retrieval Benchmark} of the DeepFashion dataset~\cite{liu2016deepfashion} is used in our experiments. 
This dataset contains $52712$ high-resolution images ($256 \times 256$) of fashion models with different clothing items in different poses.
Images of the same person in the same clothes are paired for training and testing.
We split the dataset according to the personal identity so that the identities of the training and testing sets do not overlap.
A total of $101,966$ pairs are randomly selected for training and $8,570$ pairs for testing.

\noindent
\textbf{Implementation Details.} 
In our experiments, reference features with resolutions as $32 \times 32$ and $64 \times 64$ are extracted and warped to generate the final results.
Considering the huge deformations between the reference images and target images, we train the model in stages to avoid it getting stuck in bad local minimas. 
The deformation estimation module is first pre-trained. Then we train the whole model in an end-to-end manner. 
The batch size is set to $24$ for all experiments. 
We use the historical average technique~\cite{salimans2016improved} to update the average model by weighted averaging current parameters with previous parameters. More details can be found in the \textit{Supplementary Materials}.

\begin{table*}[]
\setlength\extrarowheight{1pt}
\centering
\begin{tabular}{p{1.2cm}||P{1.5cm}P{1.5cm}P{1.5cm}P{1.5cm}P{1.5cm}P{1.5cm}P{1.5cm}}

\hline
                   & VU-Net & Def-GAN & Pose-Attn & Intr-Flow & ADGAN    & GFLA   & Ours   \\ \hline
                   &        &         &           &           &          &        &        \\ [-10pt]\hline
SSIM $\uparrow$    & 0.6738 & 0.6836  & 0.6714    & 0.6968    & 0.6736   & 0.7074 & \textbf{0.7113} \\
LPIPS $\downarrow$ & 0.2637 & 0.2330  & 0.2533    & 0.1875    & 0.2250   & 0.1962 & \textbf{0.1813} \\
FID $\downarrow$   & 23.669 & 18.460  & 20.728    & 13.014    & 14.546   & 9.9125 & \textbf{9.4502} \\
FR  $\uparrow$     & 4.12   & 14.40   &   9.56    &  16.80    & 29.08   &   18.88 & \textbf{30.00}       \\ \hline
\end{tabular}
\caption{The evaluation results compared with several state-of-the-art methods. 
SSIM and LPIPS calculate the reconstruction errors. 
FID indicates the realism of the generated images.
Fooling rate (FR) is obtained by human subjective studies. It represents the probability that the generated images are mistaken for real images.}
\label{tab:object}
\end{table*}

\subsection{Comparisons with State-of-the-arts}
We compare our model with several state-of-the-art methods including VU-Net~\cite{esser2018variational}, Def-GAN~\cite{siarohin2018deformable}, Pose-Attn\cite{zhu2019progressive}, Intr-Flow~\cite{li2019dense}, ADGAN~\cite{men2020controllable}, and GFLA~\cite{ren2020deep}.
The released weights of these methods are used for evaluation. 

\begin{figure*}[t]
\begin{center}
\includegraphics[width=0.99\linewidth]{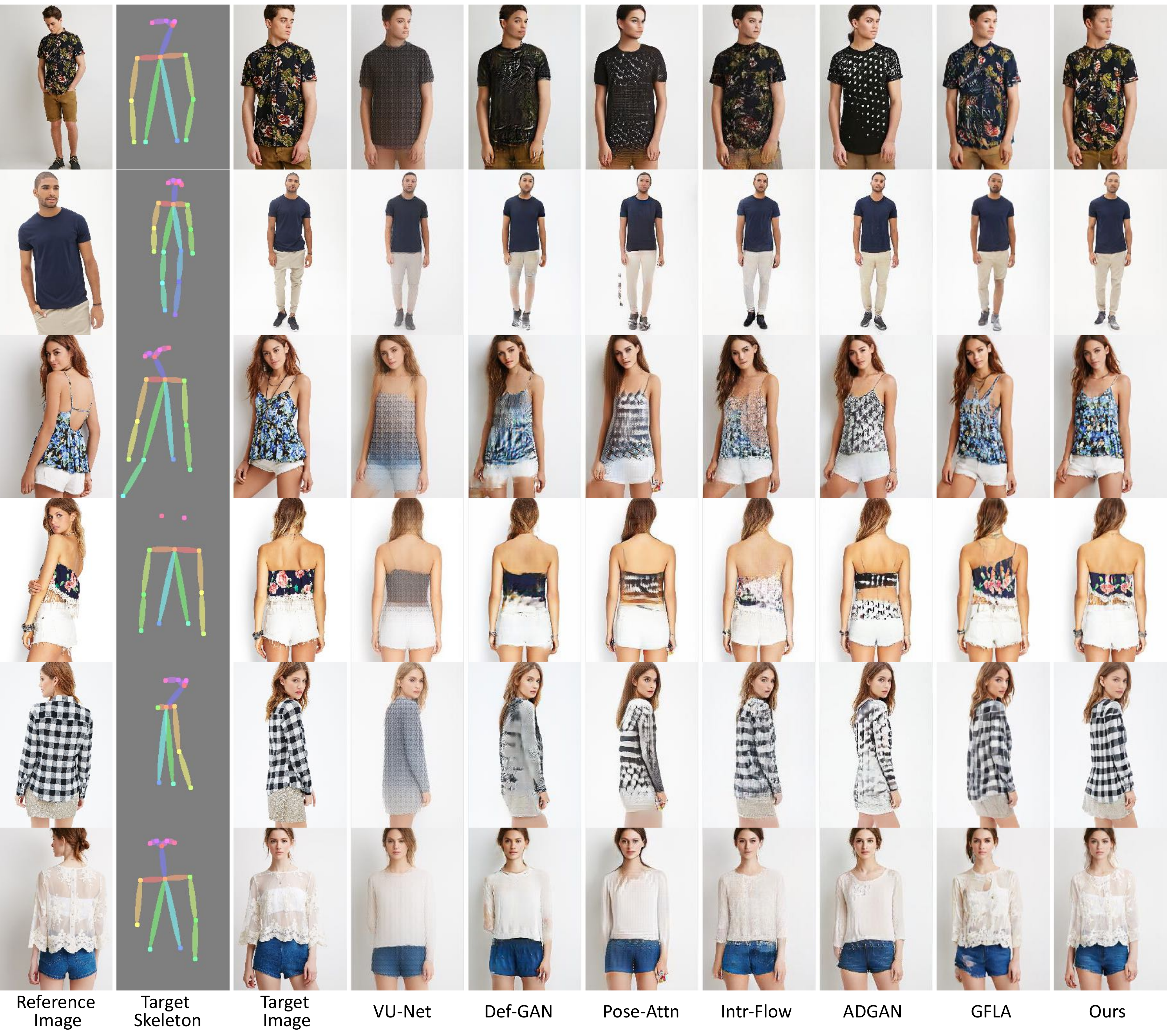}
\end{center}
\caption{The qualitative comparisons with several state-of-the-art methods including VU-Net~\cite{esser2018variational}, Def-GAN~\cite{siarohin2018deformable}, Pose-Attn\cite{zhu2019progressive}, Intr-Flow~\cite{li2019dense}, ADGAN~\cite{men2020controllable}, and GFLA~\cite{ren2020deep}}
\label{fig:comparison}
\Description{The qualitative comparisons.}
\end{figure*}

\begin{figure}[t]
\begin{center}
\includegraphics[width=0.99\linewidth]{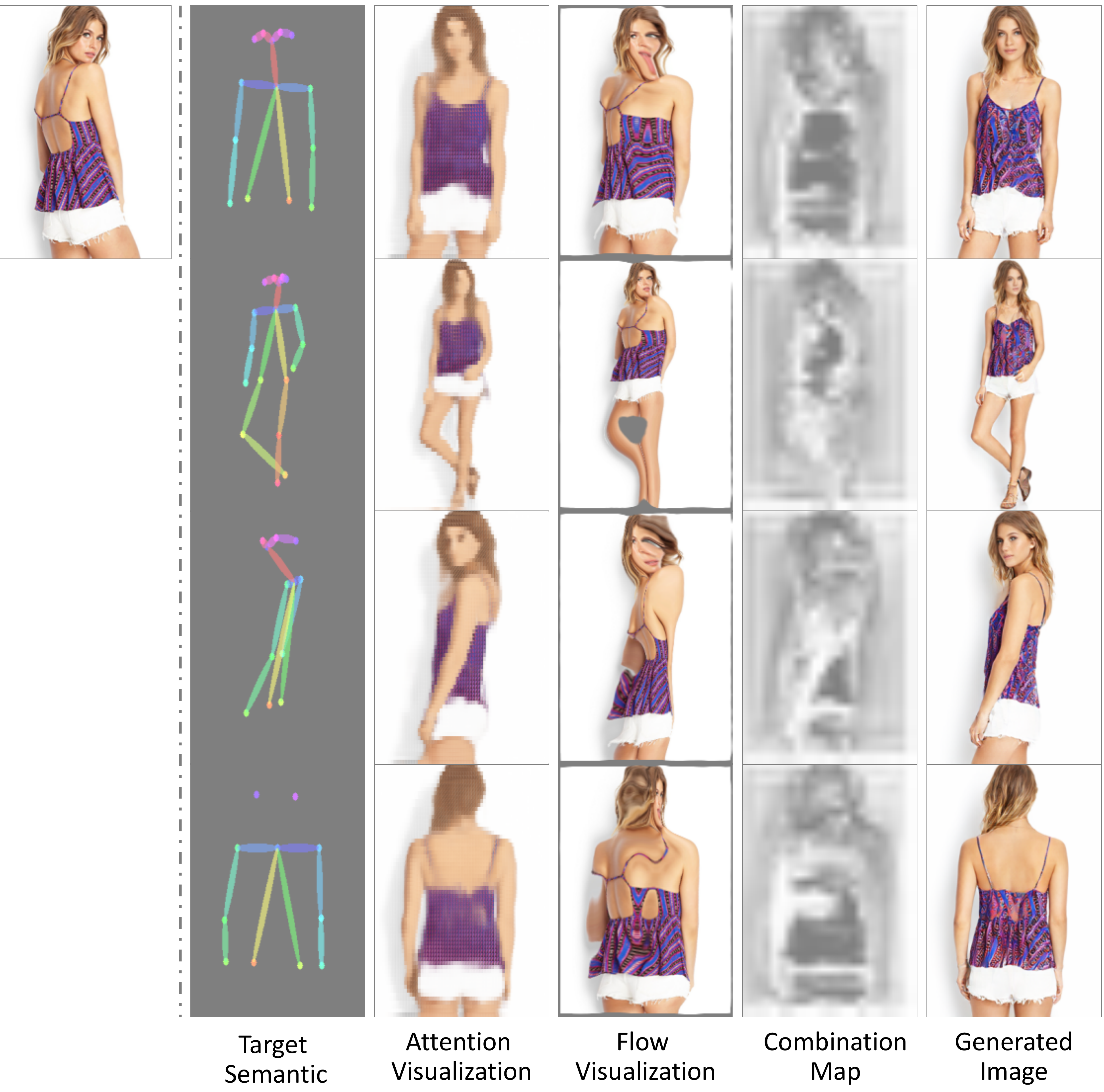}
\end{center}
\caption{More results of our model. Deformations and combination maps estimated for $64 \times 64$ feature maps are provided. 
The visualizations are resized for illustration.}
\label{fig:ours}
\Description{More results of our model.}
\end{figure}

\begin{figure}[]
\begin{center}
\includegraphics[width=1\linewidth]{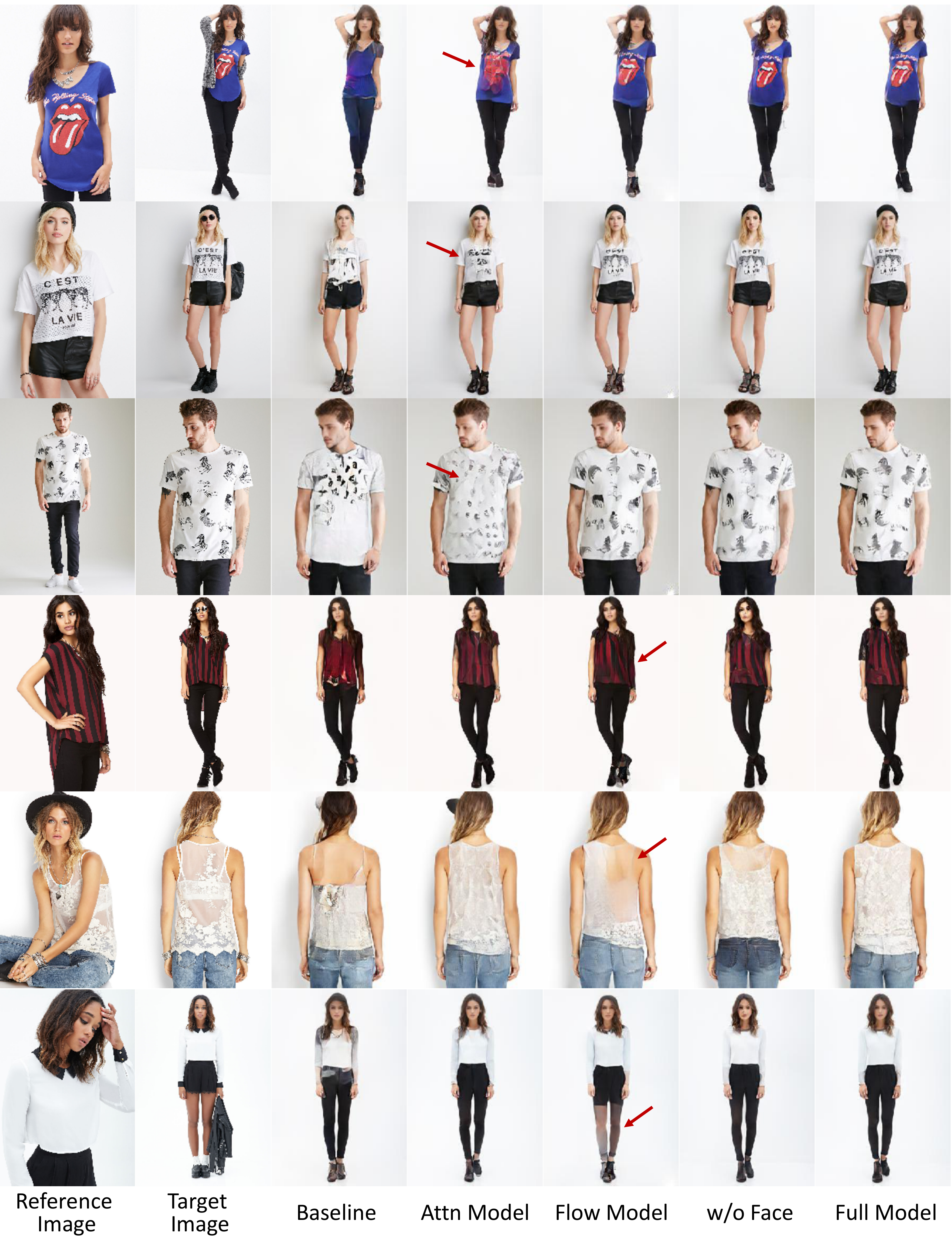}
\end{center}
\caption{The qualitative results of the ablation study. We mark some typical artifacts with red arrows.}
\label{fig:ablation}
\Description{The generated examples of the ablation models.}
\end{figure}

\noindent
\textbf{Qualitative Results.} 
We provide the typical qualitative results in Fig.~\ref{fig:comparison} for intuitive comparisons.
It can be seen that without using spatial transformation blocks, VU-Net fails to reconstruct the details of the clothes. 
Def-GAN and Pose-Attn can generate images with relatively reasonable structures. 
However, obvious artifacts can be observed in images with complex textures.
ADGAN first extracts the semantic embeddings from the reference images and then injects the extracted variables into target skeletons. 
This model can generate realistic results for solid color clothes. 
However, since the extracted 1D embeddings do not contain the image spatial information, it struggles to reconstruct complex textures.

Using the flow-based operation as the spatial transformation modules, Intr-Flow and GFLA can extract vivid neural textures from the reference images and generate results with realistic details. 
However, their results may suffer from inaccurate structures, which is more evident when complex deformations and severe occlusions are observed.  
The main reason is that the poor gradients provided by the warping operation hinder the model to estimate accurate deformation for each position.
By combining the flow-based operation with the attention operation, our model can generate images with not only accurate structures but also vivid textures.
Besides, we provide more results of our model in Fig.~\ref{fig:ours}. 
The corresponding combination maps, as well as the visualizations of the deformations are also given. 
It can be seen that the deformations estimated by the flow fields are selected to generate complex textures. 
For the smooth areas (\eg arms, legs), the deformations estimated by the attention correlation matrices are selected.
Thus, our model can use the reference information to synthesize realistic results.

\noindent
\textbf{Quantitative Results.} 
The evaluation results are shown in Tab.~\ref{tab:object}. 
We use \textit{Structure Similarity Index Measure} (SSIM)~\cite{wang2004image} and \textit{Learned Perceptual Image Patch Similarity} (LPIPS)~\cite{zhang2018unreasonable} to evaluate the reconstruction errors between the generated images and the ground-truth images. 
SSIM is a commonly used pixel-level image similarity indicator. 
However, as discussed in paper~\cite{zhang2018unreasonable}, pixel-level distance metrics may be insufficient for assessing perceptual quality.
Therefore, LPIPS is further employed to evaluate the perceptual distance by calculating feature differences using a network trained on human judgments.
It can be seen that our model achieves competitive SSIM and LPIPS scores, which means that we can generate images with better perceptual quality and fewer reconstruction errors.
Besides, \textit{Fr\'echet Inception Distance} (FID)~\cite{heusel2017gans} is used to measure the distance between the distributions of synthesized images and real images.
This metric indicates the realism of the generated images. 
Our model achieves the best FID score compared with the state-of-the-art models.

Since the objective metrics have their own limitations, their results may mismatch with the actual subjective perceptions~\cite{zhang2018unreasonable}. Therefore, we conduct a user study to compare the subjective quality. 
Image pairs of the ground-truth and generated images are shown to volunteers who are expected to select the more realistic image from each data pair. 
The \textit{Fooling Rate} (FR) is calculated as the final score of the user study.
This test is implemented on Amazon Mechanical Turk (MTurk).
We randomly select $550$ images as the test set.
Each image pair is compared $5$ times by different volunteers.
A total of $350$ volunteers participate in this experiment where each volunteer conducts $55$ comparisons.
The evaluation results are shown in Tab.~\ref{tab:object}. It can be seen that our model achieves the best FR score, which means that we can generate more realistic images.

\subsection{Ablation Study}
In this section, we evaluate the efficacy of the proposed modules by comparing our model with several variants. 

\noindent
\textbf{Baseline.}
A baseline model is trained to prove the efficacy of the deformation modules. 
An auto-encoder network is used to design this model. 
The reference images and target skeletons are concatenated as the model inputs.
We train this model using the same loss functions as that of our image generation module.

\noindent
\textbf{Attention Model.}
The attention model is used to evaluate the performance gain brought by the attention correlation estimator.
We remove the flow field estimator and use the attention correlation matrices $\mathbf{C}$ as the final deformations.

\noindent
\textbf{Flow Model.} Similarly, the flow model is designed to evaluate the efficacy of the flow field estimator. The attention correlation estimator is removed from the proposed model. We take the flow fields $\mathbf{w}$ as the final deformations to warp the reference features. 

\noindent
\textbf{Without Face Loss.} This model is used to show the efficacy of face reconstruction loss $\mathcal{L}_{face}$ on the quality of the generated images. We remove the face reconstruction loss when training this model.

\noindent
\textbf{Full Model.} The proposed model with both attention correlation estimator and flow field estimator is used in this model.

The qualitative results of the ablation models are provided in Fig.~\ref{fig:ablation}. 
It can be seen that the baseline model generates images with correct poses and identities. 
However, it cannot preserve the textures of the reference images.
By using the spatial transformation modules, both the attention model and the flow model can efficiently deform the reference images and generate targets with realistic details.
However, different types of artifacts can be observed in their results.
The powerful attention operation helps the model extract long-term correlations and generate accurate target structures. 
However, the dense connections affect the model to benefit from the image locality. 
As shown in the top three rows in Fig.~\ref{fig:ablation}, the complex textures of reference images cannot be reconstructed well. 
Meanwhile, as this operation may destroy the spatial distributions of the reference images, the special patterns (\eg logos) are not generated.
The flow-based operation can build correlations between adjacent deformations.
Thus, it helps to reconstruct the patterns and complex textures by extracting the whole local patches from reference images. 
However, as shown in the bottom three rows in Fig.~\ref{fig:ablation}, this operation fails to estimate accurate deformations for all target positions, which may lead to inaccurate target structures. Our full model benefits from both transformation modules and generates images with not only accurate structures but also realistic details. 
Meanwhile, compared with the w/o face model, the full model produces more realistic faces, which improves the perceptual quality of the final images.

\begin{table}[]
\setlength\extrarowheight{1pt}
\centering
\resizebox{0.48\textwidth}{!}{%
\begin{tabular}{p{1cm}||P{1.5cm}P{1.5cm}P{1.5cm}P{1.5cm}P{1.5cm}}
\hline
                   & Baseline   & Attn Model & Flow Model & w/o Face & Full Model \\ \hline
                   &          &            &            &          &            \\ [-10pt] \hline
SSIM $\uparrow$    & 0.6997   & 0.7119     & 0.7084     & \textbf{0.7124}   & 0.7113     \\
LPIPS $\downarrow$ & 0.2189   & 0.1852     & 0.1911     & \textbf{0.1778}   & 0.1813     \\
FID $\downarrow$   & 16.283   & 10.636     & 10.535     & 9.9336   & \textbf{9.4502}     \\ \hline
\end{tabular}%
}
\caption{The evaluation results of the ablation study.}
\label{tab:ablation}
\end{table}

The quantitative evaluation results are shown in Tab.~\ref{tab:ablation}. 
It can be seen that the baseline model fails to achieve good evaluation results. 
The main reason is that this model lacks efficient spatial transformation blocks to deform the reference information. 
Thus, the network cannot obtain aligned feature maps to synthesize the final images.
Compared with the baseline, both the attention model and the flow model achieve significant performance gains. 
This indicates that designing efficient spatial transformation modules is crucial for this task.
Meanwhile, the attention model achieves better reconstruction scores than the flow model, which confirms that the attention operation helps to generate accurate target structures. However, the poor FID score indicates that the realism of the final images is degraded since it cannot preserve vivid reference details.
By combining the attention operation with the flow operation, both the w/o face model and the full model obtain additional performance improvements. 
Such combination helps the models take the complementary advantages of the deformation blocks.
Although the w/o face model achieves better LPIPS and SSIM scores, as shown in Fig.~\ref{fig:ablation}, due to the acuteness of the human visual system towards the faces, a dramatic improvement in visual quality is obtained by using the face reconstruction loss $\mathcal{L}_{face}$.

\begin{figure}[]
\begin{center}
\includegraphics[width=1\linewidth]{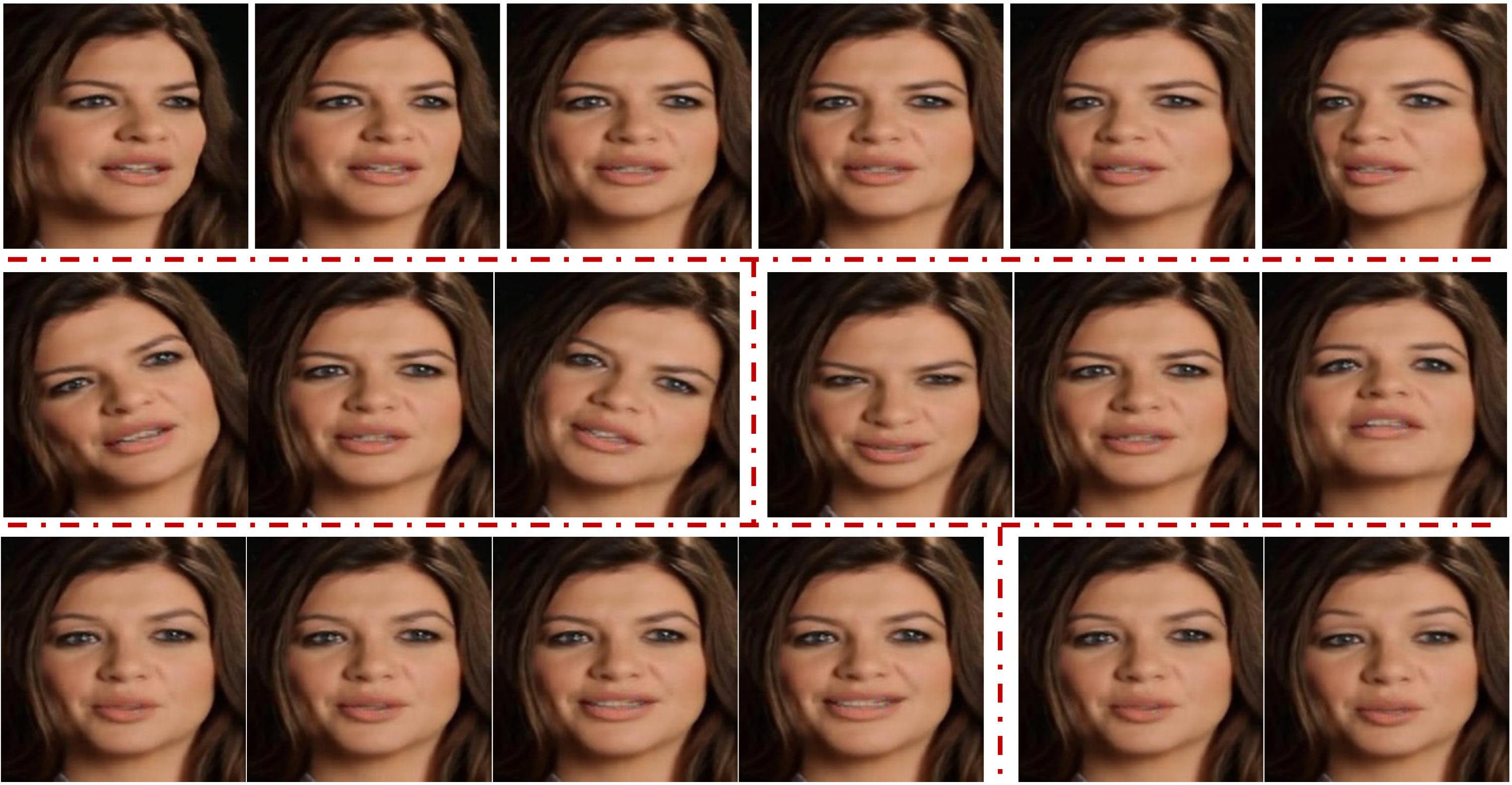}
\end{center}
\caption{The results of the portrait image editing task. The top two rows show the results of pose editing. The bottom row shows the results of expression editing.}
\label{fig:face}
\Description{The face editing results.}
\end{figure}

\subsection{Portrait Image Editing}
In this subsection, we further apply our model to tackle the portrait image editing task. 
To achieve intuitive portrait image control, we employ the three-dimensional morphable face models~\cite{blanz1999morphable,paysan20093d} (3DMMs) to describe the motions of the faces. 3DMMs allow users to control the 3D face meshes with fully disentangled semantic parameters (\eg shape, pose). The VoxCeleb dataset~\cite{nagrani2017voxceleb} which contains $22496$ talking-head videos is used for training and testing. 
The face reconstruction model proposed in paper~\cite{deng2019accurate} is employed to extract the corresponding 3DMM parameters from the images. 
The face landmarks are rendered from the extracted 3DMM parameters and used as the semantics $\mathbf{s}_r$, $\mathbf{s}_t$ of the face images $\mathbf{x}_r$ and $\mathbf{x}_t$.
Images from the same video are randomly paired as the reference and target images.
After training, the users can control the motions of portrait images by providing specific 3DMM parameters. 
Example results are shown in Fig.~\ref{fig:face}. 
Our model enables intuitively editing the poses and expressions of a given portrait image, which will find a large variety of applications in the industries such as social media, virtual reality.
Meanwhile, our model can generate images with realistic details and accurate motions. Please find more details in the \textit{Supplementary Materials}.


\section{Conclusion}
We have proposed a novel model for the pose-guided person image synthesis task by combining the attention operation with the flow-based operation. 
We empirically demonstrate the advantages and disadvantages of these two operations for the spatial transformation.
Ablation studies prove that our model can take the complementary advantages to generate images with not only accurate global structures but also realistic local details. 
Both subjective and objective experiments show the superiority of the proposed model compared with the state-of-the-arts. 
Besides, additional experiment on the portrait image editing task proves that the proposed combination can be flexibly applied to different types of data.




\bibliographystyle{ACM-Reference-Format}
\balance
\bibliography{sample-base}










\end{document}